\documentclass[10pt,twocolumn,letterpaper]{article}

\usepackage{cvpr}
\usepackage{times}
\usepackage{epsfig}
\usepackage{graphicx}
\usepackage{amsmath}
\usepackage{amssymb}

\usepackage{caption}
\usepackage{subcaption}

\usepackage[pagebackref=true,breaklinks=true,colorlinks,bookmarks=false]{hyperref}

\cvprfinalcopy 

\def\cvprPaperID{6886} 

\ifcvprfinal\fi
\begin{document}

\title{Controllable Person Image Synthesis with Attribute-Decomposed GAN}

\author{Yifang Men$^1$, Yiming Mao$^2$, Yuning Jiang$^2$, Wei-Ying Ma$^2$, Zhouhui Lian$^1$\thanks{Corresponding author. E-mail: lianzhouhui@pku.edu.cn}\\
$^1$Wangxuan Institute of Computer Technology, Peking University, China\\
$^2$Bytedance AI Lab\\}


\maketitle
\thispagestyle{empty}

\begin{abstract}
This paper introduces the Attribute-Decomposed GAN, a novel generative model for controllable person image synthesis, which can produce realistic person images with desired human attributes (e.g., pose, head, upper clothes and pants) provided in various source inputs. The core idea of the proposed model is to embed human attributes into the latent space as independent codes and thus achieve flexible and continuous control of attributes via mixing and interpolation operations in explicit style representations. Specifically, a new architecture consisting of two encoding pathways with style block connections is proposed to decompose the original hard mapping into multiple more accessible subtasks. In source pathway, we further extract component layouts with an off-the-shelf human parser and feed them into a shared global texture encoder for decomposed latent codes. This strategy allows for the synthesis of more realistic output images and automatic separation of un-annotated attributes. Experimental results demonstrate the proposed method's superiority over the state of the art in pose transfer and its effectiveness in the brand-new task of component attribute transfer.

\end{abstract}
\vspace{-15pt}
\section{Introduction}

\begin{figure}
\setlength{\abovecaptionskip}{0cm}
\setlength{\belowcaptionskip}{-0.4cm}
\includegraphics[width=1\linewidth]{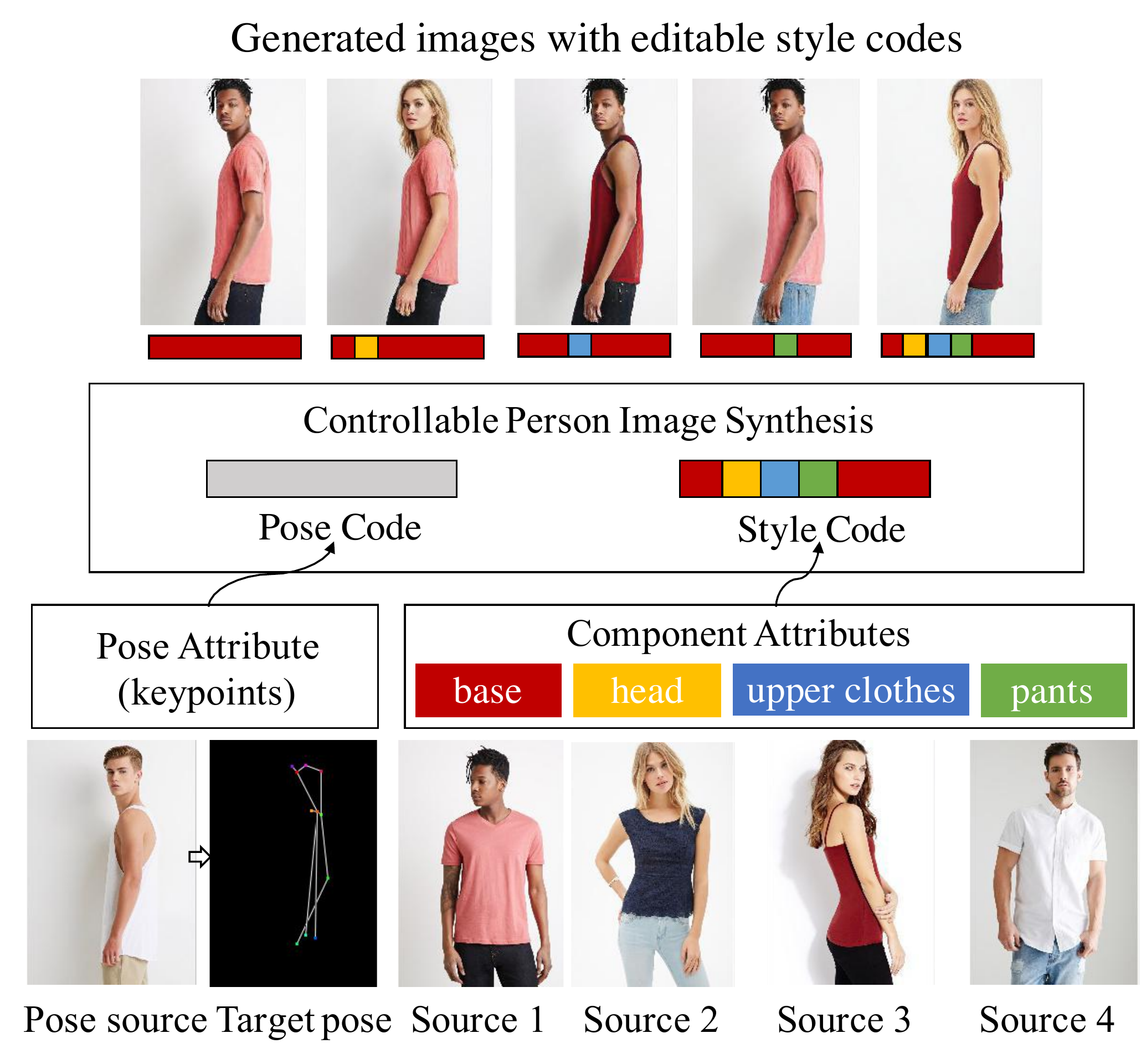}
\caption{Controllable person image synthesis with desired human attributes provided by multiple source images. Human attributes including pose and component attributes are embedded into the latent space as the pose code and decomposed style code. Target person images can be generated in user control with the editable style code. }
\label{fig:intro}
\end{figure}

Person image synthesis (PIS), a challenging problem in areas of Computer Vision and Computer Graphics, has huge potential applications for image editing, movie making, person re-identification (Re-ID), virtual clothes try-on and so on. An essential task of this topic is pose-guided image generation~\cite{ma2017pose,ma2018disentangled,esser2018variational,siarohin2018deformable}, rendering the photo-realistic images of people in arbitrary poses, which has become a new hot topic in the community. Actually, not only poses but also many other valuable human attributes can be used to guide the synthesis process.

In this paper, we propose a brand-new task that aims at synthesizing person images with controllable human attributes, including pose and component attributes such as head, upper clothes and pants. As depicted in Figure~\ref{fig:intro}, users are allowed to input multiple source person images to provide desired human attributes respectively. The proposed model embeds component attributes into the latent space to construct the style code and encodes the keypoints-based 2D skeleton extracted from the person image as the pose code, which enables intuitive component-specific (pose) control of the synthesis by freely editing the style (pose) code. Thus, our method can automatically synthesize high-quality person images in desired component attributes under arbitrary poses and can be widely applied in not only pose transfer and Re-ID, but also garment transfer and attribute-specific data augmentation (e.g., clothes commodity retrieval and recognition).

Due to the insufficiency of annotation for human attributes, the simplicity of keypoint representation and the diversity of person appearances, it is challenging to achieve the goal mentioned above using existing methods. Pose transfer methods firstly proposed by~\cite{ma2017pose} and later extended by~\cite{ma2018disentangled,esser2018variational,siarohin2018deformable,zhu2019progressive} mainly focus on pose-guided person image synthesis and they do not provide user control of human attributes such as head, pants and upper clothes. Moreover, because of the non-rigid nature of human body, it is difficult to directly transform the spatially misaligned body-parts via convolution neural networks and thus these methods are unable to produce satisfactory results. Appearance transfer methods~\cite{zanfir2018human,yang2016detailed,pons2017clothcap} allow users to transfer clothes from one person to another by estimating a complicated 3D human mesh and warping the textures to fit for the body topology. Yet, these methods fail to model the intricate interplay of the inherent shape and appearance, and lead to unrealistic results with deformed textures. Another type of appearance transfer methods~\cite{raj2018swapnet,lassner2017generative,zhu2017your} try to model clothing textures by feeding the entire source person image into neural networks, but they cannot transfer human attributes from multiple source person images and lack the capability of component-level clothing editing.

The notion of attribute editing is commonly used in the field of facial attribute manipulation~\cite{he2019attgan,zhang2018generative,yin2017semi}, but to the best of our knowledge this work is the first to achieve attribute editing in the task of person image synthesis. Different from pervious facial attribute editing methods which require strict attribute annotation (e.g., smiling, beard and eyeglasses exist or not in the training dataset), the proposed method does not need any annotation of component attributes and enables automatic and unsupervised attribute separation via delicately-designed modules. In another aspect, our model is trained with only a partial observation of the person and needs to infer the unobserved body parts to synthesize images in different poses and views. It is more challenging than motion imitation methods~\cite{chan2019everybody,aberman2019learning,wang2018video}, which utilize all characters performing a series of same motions to disentangle the appearance and pose, or train one model for each character by learning a mapping from 2D pose to one specific domain.

To address the aforementioned challenges, we propose a novel controllable person image synthesis method via an Attribute-Decomposed GAN. In contrast to previous works~\cite{ma2017pose,balakrishnan2018synthesizing,siarohin2018deformable} forcedly learn a mapping from concatenated conditions to the target image, we introduce a new architecture of generator with two independent pathways, one for pose encoding and the other for decomposed component encoding. For the latter, our model first separates component attributes automatically from the source person image via its semantic layouts which are extracted with a pretrained human parser. Component layouts are fed into a global texture encoder with multi-branch embeddings and their latent codes are recombined in a specific order to construct the style code. Then the cascaded style blocks, acting as a connection of two pathways, inject the component attributes represented by the style code into the pose code by controlling the affine transform parameters of AdaIN layer. Eventually, the desired image can be reconstructed from target features. In summary, our contributions are threefold:

\begin{itemize}
\item We propose a brand-new task that synthesizes person images with controllable human attributes by directly providing different source person images, and solve it by modeling the intricate interplay of the inherent pose and component-level attributes.
\item We introduce the Attribute-Decomposed GAN, a neat and effective model achieving not only flexible and continuous user control of human attributes, but also a significant quality boost for the original PIS task.
\item We tackle the challenge of insufficient annotation for human attributes by utilizing an off-the-shelf human parser to extract component layouts, making an automatic separation of component attributes.
\end{itemize}

\begin{figure*}
\begin{center}
\setlength{\abovecaptionskip}{0cm}
\setlength{\belowcaptionskip}{-0.5cm}
\includegraphics[width=0.9\linewidth]{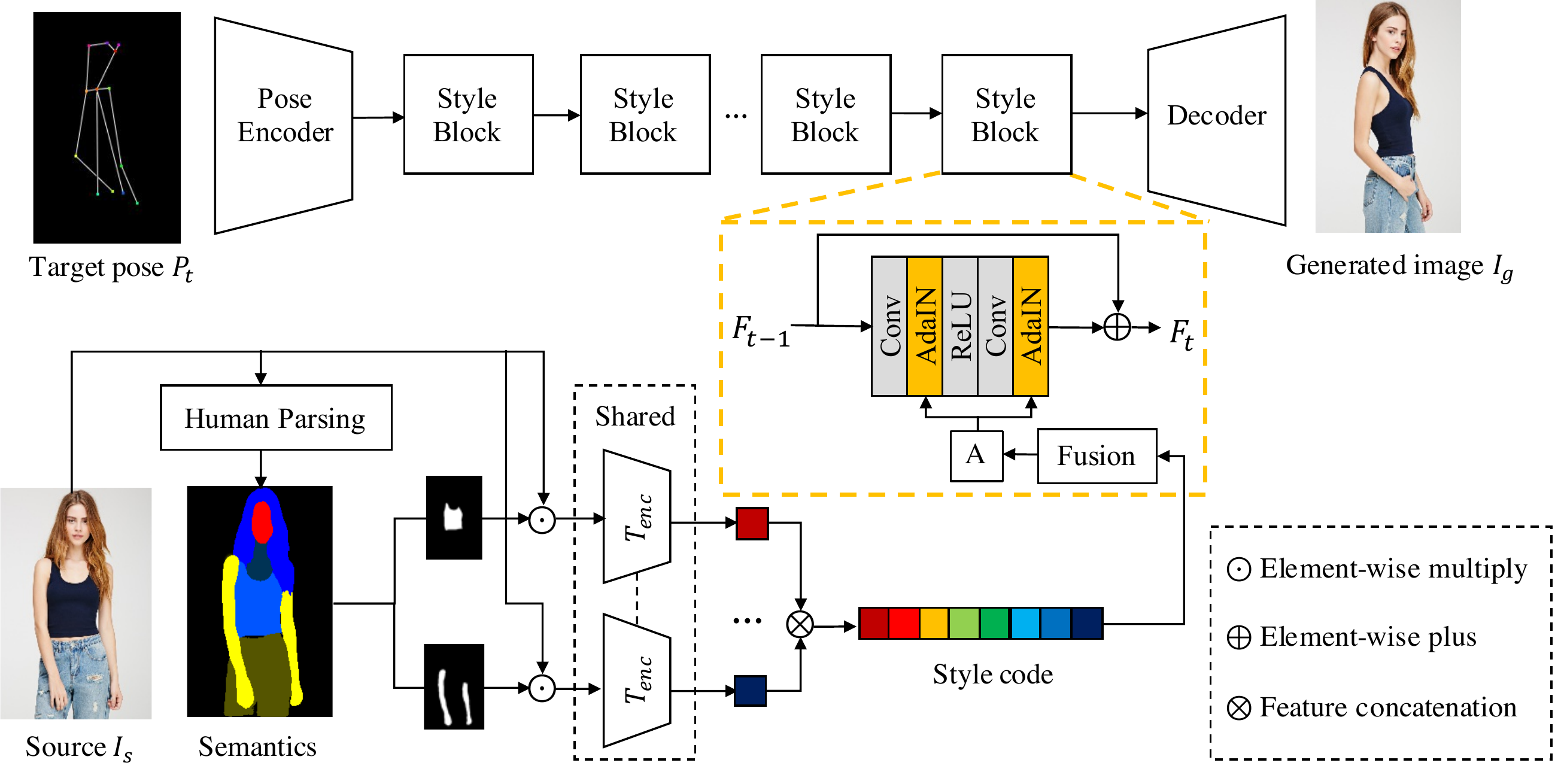}
\caption{An overview of the network architecture of our generator. The target pose and source person are embedded into the latent space via two independent pathways, called pose encoding and decomposed component encoding, respectively. For the latter, we employ a human parser to separate component attributes and encode them via a global texture encoder. A series of style blocks equipped with a fusion module are introduced to inject the texture style of source person into the pose code by controlling the affine transform parameters in AdaIN layers. Finally, the desired image is reconstructed via a decoder.}
\label{fig:generator}
\end{center}
\end{figure*}

\vspace{-5pt}
\section{Related Work}
\subsection{Image Synthesis}
Due to their remarkable results, Generative Adversarial Networks (GANs)~\cite{goodfellow2014generative} have become powerful generative models for image synthesis~\cite{isola2017image,zhu2017unpaired,brock2018large} in the last few years. The image-to-image translation task was solved with conditional GANs~\cite{mirza2014conditional} in Pix2pix~\cite{isola2017image} and extended to high-resolution level in Pix2pixHD~\cite{wang2018high}. Zhu et al.~\cite{zhu2017unpaired} introduced an unsupervised method, CycleGAN, exploiting cycle consistency to generate the image from two domains with unlabeled images. Much of the work focused on improving the quality of GAN-synthesized images by stacked architectures~\cite{zhang2017stackgan,mordido2018dropout}, more interpretable latent representations~\cite{chen2016infogan} or self-attention mechanism~\cite{zhang2018self}. StyleGAN ~\cite{karras2019style} synthesized impressive images by proposing a brand-new generator architecture which controls generator via the adaptive instance normalization (AdaIN)~\cite{huang2017arbitrary}, the outcome of style transfer literature~\cite{gatys2015texture,gatys2016image,johnson2016perceptual}. However, these techniques have limited scalability in handling attributed-guided person synthesis, due to complex appearances and simple poses with only several keypoints. Our method built on GANs overcomes these challenges by a novel generator architecture designed with attribute decomposition.

\subsection{Person Image Synthesis}
Up to now, many techniques have been proposed to synthesize person images in arbitrary poses using adversarial learning. PG\textsuperscript{2}~\cite{ma2017pose} firstly proposed a two-stage GAN architecture to generate person images, in which the person with the target pose is coarsely synthesized in the first stage, and then refined in the second stage. Esser et al.~\cite{esser2018variational} leveraged a variational autoencoder combined with the conditional U-Net~\cite{ronneberger2015u}to model the inherent shape and appearance. Siarohin et al.~\cite{siarohin2018deformable} used a U-Net based generator with deformable skip connections to alleviate the pixel-to-pixel misalignments caused by pose differences. A later work by Zhu et al.~\cite{zhu2019progressive} introduced cascaded Pose-Attentional Transfer Blocks into generator to guide the deformable transfer process progressively.~\cite{pumarola2018unsupervised,song2019unsupervised} utilized a bidirectional strategy for synthesizing person images in an unsupervised manner. However, these methods only focused on transferring the pose of target image to the reference person and our method achieved a controllable person image synthesis with not only pose guided, but also component attributes (e.g., head, upper clothes and pants) controlled. Moreover, more realistic person images with textural coherence and identical consistency can be produced.

%

\section{Method Description}

Our goal is to synthesize high-quality person images with user-controlled human attributes, such as pose, head, upper clothes and pants. Different from previous attribute editing methods~\cite{he2019attgan,yin2017semi,zhang2018generative} requiring labeled data with binary annotation for each attribute, our model achieves automatic and unsupervised separation of component attributes by introducing a well-designed generator. Thus, we only need the dataset that contains person images $\{I\in R^{3\times H\times W}\}$ with each person in several poses. The corresponding keypoint-based pose $P\in R^{18\times H\times W}$ of $I$, 18 channel heat map that encodes the locations of 18 joints of a human body, can be automatically extracted via an existing pose estimation method~\cite{cao2017realtime}. During training, a target pose $P_t$ and a source person image $I_s$ are fed into the generator and a synthesized image $I_g$ following the appearance of $I_s$ but under the pose $P_t$ will be challenged for realness by the discriminators. In the following, we will give a detailed description for each part of our model.

\subsection{Generator}
Figure~\ref{fig:generator} shows the architecture of our generator, whose inputs are the target pose $P_t$ and source person image $I_s$, and the output is the generated image $I_g$ with source person $I_s$ in the target pose $P_t$. Unlike the generator in ~\cite{ma2017pose} which directly concatenates the source image and target condition together as input to a U-Net architecture and forcedly learns a result under the supervision of the target image $I_t$, our generator embeds the target pose $P_t$ and source person $I_s$ into two latent codes via two independent pathways, called pose encoding and decomposed component encoding, respectively. These two pathways are connected by a series of style blocks, which inject the texture style of source person into the pose feature. Finally, the desired person image $I_g$ is reconstructed from target features by a decoder.

\vspace{-10pt}
\subsubsection{Pose encoding}

In the pose pathway, the target pose $P_t$ is embedded into the latent space as the pose code $C_{pose}$ by a pose encoder, which consists of $N$ down-sampling convolutional layers ($N=2$ in our case), following the regular configuration of encoder.

\begin{figure}
\setlength{\abovecaptionskip}{0cm}
\setlength{\belowcaptionskip}{-0.2cm}
\includegraphics[width=1\linewidth]{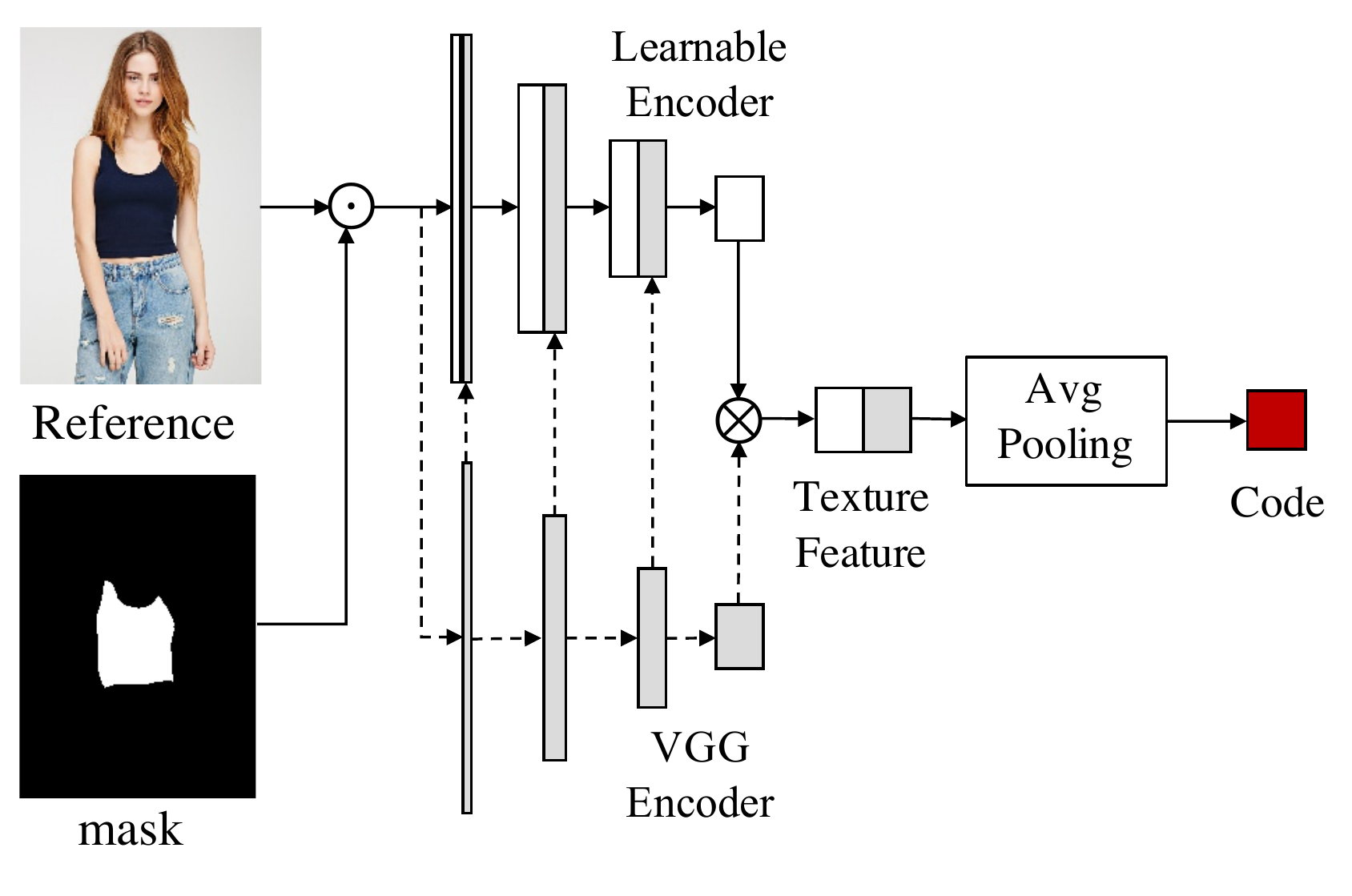}
\caption{Details of the texture encoder in our generator. A global texture encoding is introduced by concatenating the output of learnable encoder and fixed VGG encoder.}
\label{fig:texture_enc}
\end{figure}

\subsubsection{Decomposed component encoding}
In the source pathway, the source person image $I_s$ is embedded into the latent space as the style code $C_{sty}$ via a module called decomposed component encoding (DCE). As depicted in Figure~\ref{fig:generator}, this module first extracts the semantic map $S$ of source person $I_s$ with an existing human parser~\cite{gong2017look} and converts $S$ into a $K$-channel heat map $M \in R^{K\times H\times W}$. For each channel $i$, there is a binary mask $M_i \in R^{H\times W}$ for the corresponding component (e.g., upper clothes). The decomposed person image with component $i$ is computed by multiplying the source person image with the component mask $M_i$
\vspace{-5 pt}
\begin{equation}
I_s^i = I_s \odot M_i,\vspace{-5 pt}
\end{equation}
where $\odot$ denotes element-wise product. $I_s^i$ is then fed into the texture encoder $T_{enc}$ to acquire the corresponding style code $C_{sty}^i$ in each branch by
\vspace{-4 pt}
\begin{equation}
C_{sty}^i = T_{enc}(I_s^i),\vspace{-4 pt}
\end{equation}
where the texture encode $T_{enc}$ is shared for all branches and its detailed architecture will be described below. Then all $C_{sty}^i, i=1…K$ will be concatenated together in a top-down manner to get the full style code $C_{sty}$.

\begin{figure}
\setlength{\abovecaptionskip}{0cm}
\setlength{\belowcaptionskip}{-0.2cm}
\includegraphics[width=1\linewidth]{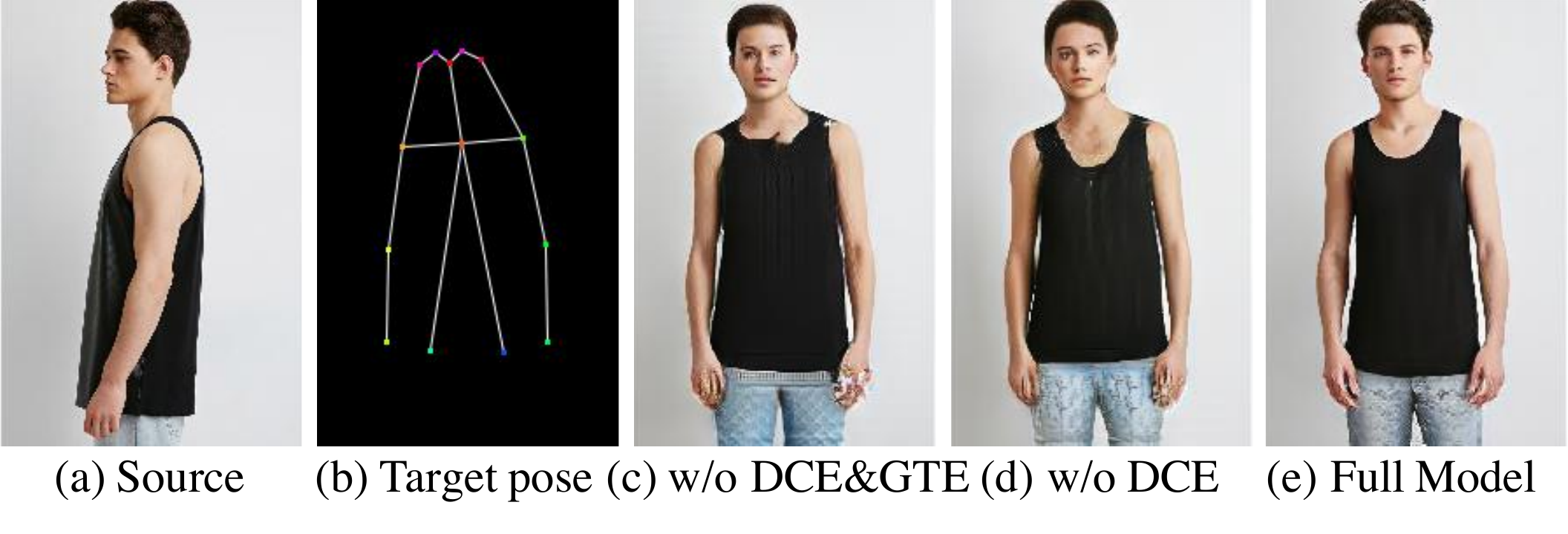}
\caption{Visualization effects of the DCE and GTE. (a) A source person and (b) a target pose for inputs. (c) The result generated without either DCE or GTE. (d) The result generated without only DCE. (e) The result generated with both two modules. }
\label{fig:eff_dce_gte}
\end{figure}

\begin{figure}
\setlength{\abovecaptionskip}{0cm}
\setlength{\belowcaptionskip}{-0.3cm}
\includegraphics[width=1\linewidth]{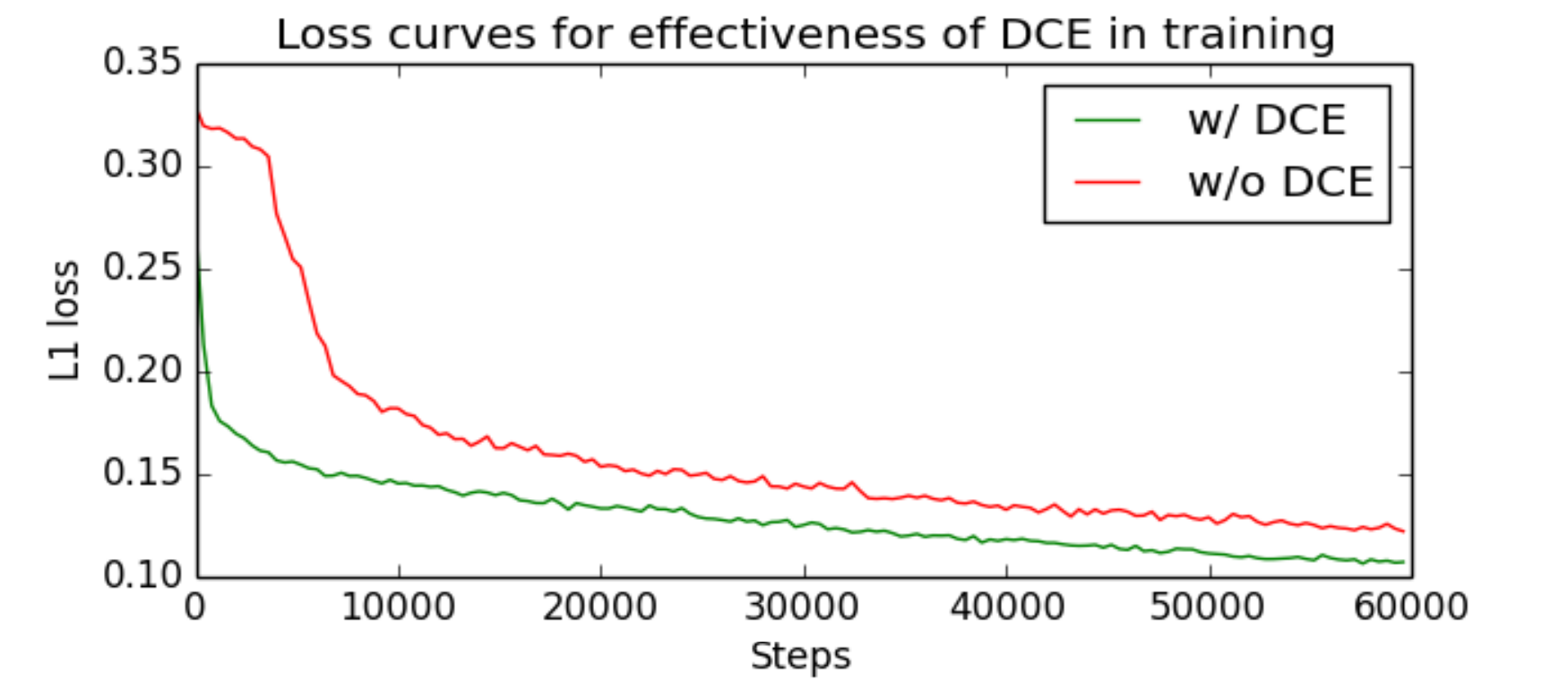}
\caption{Loss curves for the effectiveness of our DCE module in the training process.}
\label{fig:loss}
\end{figure}

In contrast to the common solution that directly encodes the entire source person image, this intuitive DCE module decomposes the source person into multiple components and recombines their latent codes to construct the full style code. Such an intuitive strategy kills two birds with one stone: 1) It speeds up the convergence of model and achieves more realistic results in less time. Due to the complex structure of the manifold that is constituted of various person images with different clothes and poses, it is hard to encode the entire person with detailed textures, but much simpler to only learn the features of one component of the person. Also, different components can share the same network parameters for color encoding and thus DCE implicitly provides a data augmentation for texture learning. The loss curves for the effects of our DCE module in training are shown in Figure~\ref{fig:loss} and the visualization effects are provided in Figure~\ref{fig:eff_dce_gte} (d)(e). 2) It achieves an automatic and unsupervised attribute separation without any annotation in the training dataset, which utilizes an existing human parser for spatial decomposition. Specific attributes are learned in the fixed positions of the style code. Thus we can easily control component attributes by mixing desired component codes extracted from different source persons.

For the texture encoder, inspired by a style transfer method~\cite{huang2017arbitrary} which directly extracts the image code via a pretrained VGG network to improve the generalization ability of texture encoding, we introduce an architecture of global texture encoding by concatenating the VGG features in corresponding layers to our original encoder, as shown in Figure~\ref{fig:texture_enc}. The values of parameters in the original encoder are learnable while those in the VGG encoder are fixed. Since the fixed VGG network is pretrained on the COCO dataset~\cite{lin2014microsoft} and it has seen many images with various textures, it has a global property and strong generalization ability for in-the-wild textures. But unlike the typical style transfer task~\cite{huang2017arbitrary,gatys2016image} requiring only a roughly reasonable result without tight constraints, our model needs to output the explicitly specified result for a given source person in the target pose. It is difficult for the network with a fixed encoder to fit such a complex model and thus the learnable encoder is introduced, combined with the fixed one. The effects of the global texture encoding (GTE) are shown in Figure~\ref{fig:eff_dce_gte} (c)(d).

\begin{figure}
\setlength{\abovecaptionskip}{0cm}
\setlength{\belowcaptionskip}{-0.5cm}
\includegraphics[width=1\linewidth]{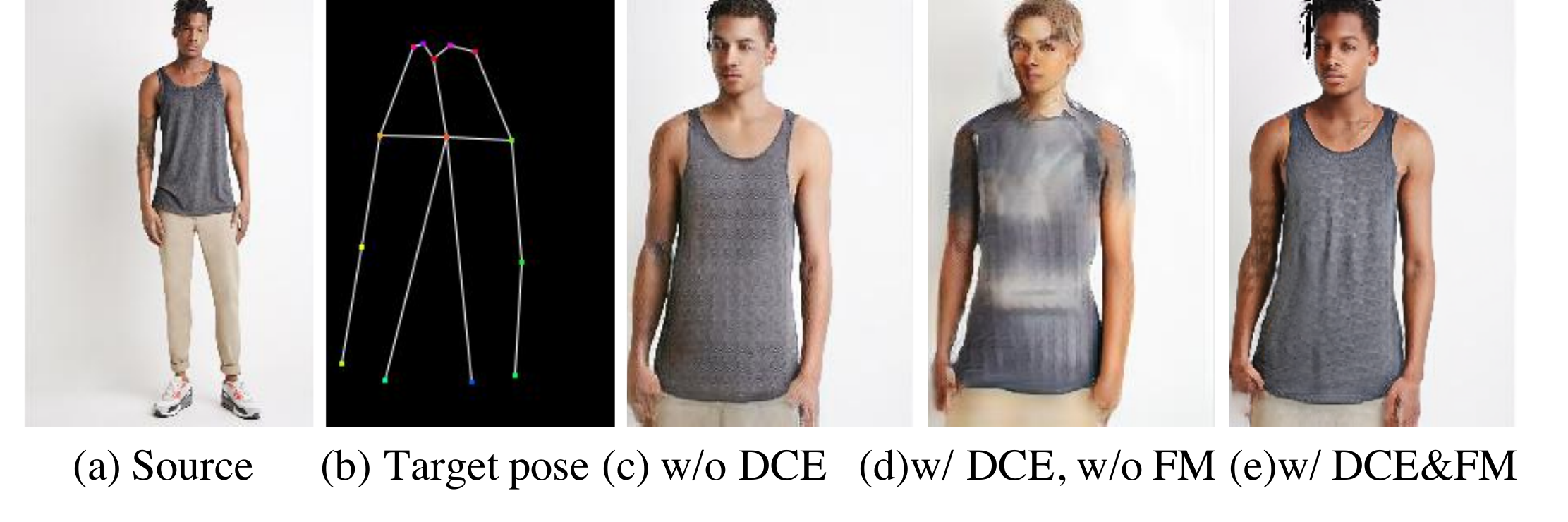}
\caption{Auxiliary effects of the fusion module (FM) for DCE. (a) A source person and (b) a target pose for inputs. (c) The result generated without DCE. (d) The result generated with DCE introduced but no FM contained in style blocks. (e) The result generated with both DCE and FM.}
\label{fig:eff_fm}
\end{figure}

\subsubsection{	Texture style transfer }
Texture style transfer aims to inject the texture pattern of source person into the feature of target pose, acting as a connection of the pose code and style code in two pathways. This transfer network consists of several cascaded style blocks, each one of which is constructed by a fusion module and residual conv-blocks equipped with AdaIN. For the $t^{th}$ style block, its inputs are deep features $F_{t-1}$ at the output of the previous block and the style code $C_{sty}$. The output of this block can be computed by

\vspace{-10 pt}

\begin{equation}
F_{t} =\varphi_t(F_{t-1}, A) + F_{t-1},
\end{equation}
where $F_{t-1}$ firstly goes through conv-blocks $\varphi_t$, whose output is added back to $F_{t-1}$ to get the output $F_t$, $F_0 = C_{pose}$ in the first block and 8 style blocks are adopted totally. $A$ denotes learned affine transform parameters (scale $\mu$ and shift $\sigma$) required in the AdaIN layer and can be used to normalize the features into the desired style~\cite{dumoulin2016learned,huang2017arbitrary}. Those parameters are extracted from the style code $C_{sty}$ via a fusion module (FM), which is an important auxiliary module for DCE. Because component codes are concatenated in a specified order to construct the style code, making a high correlation between the position and component features, this imposes much human ruled intervention and leads to a conflict with the learning tendency of the network itself. Thus we introduce FM consisting of 3 fully connected layers with the first two allowing the network to flexibly select the desired features via linear recombination and the last one providing parameters in the required dimensionality. FM can effectively disentangle features and avoid conflicts between forward operation and backward feedback. The effects of FM are shown in Figure~\ref{fig:eff_fm}. When DCE is applied to our model without FM, the result (see Figure~\ref{fig:eff_fm} (d)) is even worse than that without DCE (see Figure~\ref{fig:eff_fm} (c)). The fusion module makes our model more flexible and guarantees the proper performance of DCE.

\subsubsection{	Person image reconstruction}
With the final target features $F_{T-1}$ at the output of the last style block, the decoder generates the final image $I_g$ from $F_{T-1}$ via $N$ deconvolutional layers, following the regular decoder configuration.

\subsection{Discriminators}
Following Zhu et al.~\cite{zhu2019progressive}, we adapt two discriminators ${D}_p$ and ${D}_t$, where ${D}_p$ is used to guarantee the alignment of the pose of generated image $I_g$ with the target pose $P_t$ and ${D}_t$ is used to ensure the similarity of the appearance texture of $I_g$ with the source person $I_s$. For ${D}_p$, the target pose $P_t$ concatenated with the generated image $I_g$ (real target image $I_t$) is fed into ${D}_p$ as a fake (real) pair. For ${D}_t$, the source person image $I_s$ concatenated with $I_g$ ($I_t$) is fed into ${D}_t$ as a fake (real) pair. Both ${D}_p$ and ${D}_t$ are implemented as PatchGAN and more details can be found in ~\cite{isola2017image}.

\begin{figure*}
\setlength{\abovecaptionskip}{0.1cm}
\setlength{\belowcaptionskip}{-0.3cm}
\includegraphics[width=1\linewidth]{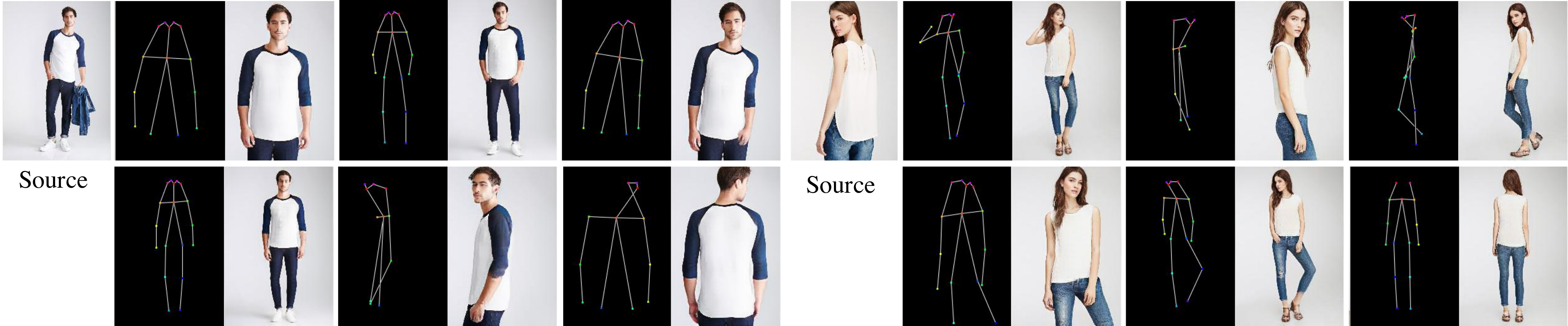}
\caption{Results of synthesizing person images in arbitrary poses.}
\label{fig:res_pose}
\end{figure*}

\subsection{Training}
Our full training loss is comprised of an adversarial term, a reconstruction term, a perceptual term and a contextual term
\vspace{-5 pt}
\begin{equation}
\mathcal{L}_{total} = \mathcal{L}_{adv}+\lambda_{rec}\mathcal{L}_{rec}+\lambda_{per}\mathcal{L}_{per}+\lambda_{CX}\mathcal{L}_{CX},
\end{equation}
where $\lambda_{rec}, \lambda_{per}$ and $\lambda_{CX}$ denote the weights of corresponding losses, respectively.

{ \noindent \bf Adversarial loss.}
We employ an adversarial loss $\mathcal{L}_{adv}$ with discriminators ${D}_p$ and ${D}_t$ to help the generator ${G}$ synthesize the target person image with visual textures similar to the reference one, as well as following the target pose. It penalizes for the distance between the distribution of real pairs ($I_s(P_t), I_t$) and the distribution of fake pairs ($I_s(P_t), I_g$) containing generated images
\begin{equation}
\begin{split}
L_{adv}=&\mathbb{E}_{I_s,P_t,I_t}[log(D_t(I_s,I_t) \cdot D_p({P_t,I_t})) ]+ \\
&\mathbb{E}_{I_s,P_t}[log((1-D_t(I_s,{G}(I_s,P_t)))\\
&\cdot(1-D_p(P_t,{G}(I_s,P_t))))].
\end{split}
\end{equation}

{ \noindent \bf Reconstruction loss.}
The reconstruction loss is used to directly guide the visual appearance of the generated image similar to that of the target image $I_t$, which can avoid obvious color distortions and accelerate the convergence process to acquire satisfactory results. $\mathcal{L}_{rec}$ is formulated as the L1 distance between the generated image and target image $I_t$
\begin{equation}
L_{rec} = || {G}(I_s,P_t)-I_t ||_1.
\end{equation}

{ \noindent \bf Perceptual loss.}
Except for low-level constraints in the RGB space, we also exploit deep features extracted from certain layers of the pretrained VGG network for texture matching, which has been proven to be effective in image synthesis~\cite{esser2018variational,siarohin2018deformable} tasks. The preceputal loss is computed as~\cite{zhu2019progressive}
\begin{equation}
\mathcal{L}_{per} = \frac{1}{W_{l}H_{l}C_{l}}\sum_{x=1}^{W_l}\sum_{y=1}^{H_l}\sum_{z=1}^{C_l}\parallel \phi_l(I_g)_{x,y,z}-\phi_l(I_t)_{x,y,z}  \parallel _1,
\end{equation}
where $\phi_l$ is the output feature from layer $l$ of VGG19 network, and $W_{l}, H_{l}, C_{l}$ are spatial width, height and depth of feature $\phi_l$.

\begin{figure}
\setlength{\abovecaptionskip}{-0.1cm}
\setlength{\belowcaptionskip}{-0.3cm}
\includegraphics[width=1\linewidth]{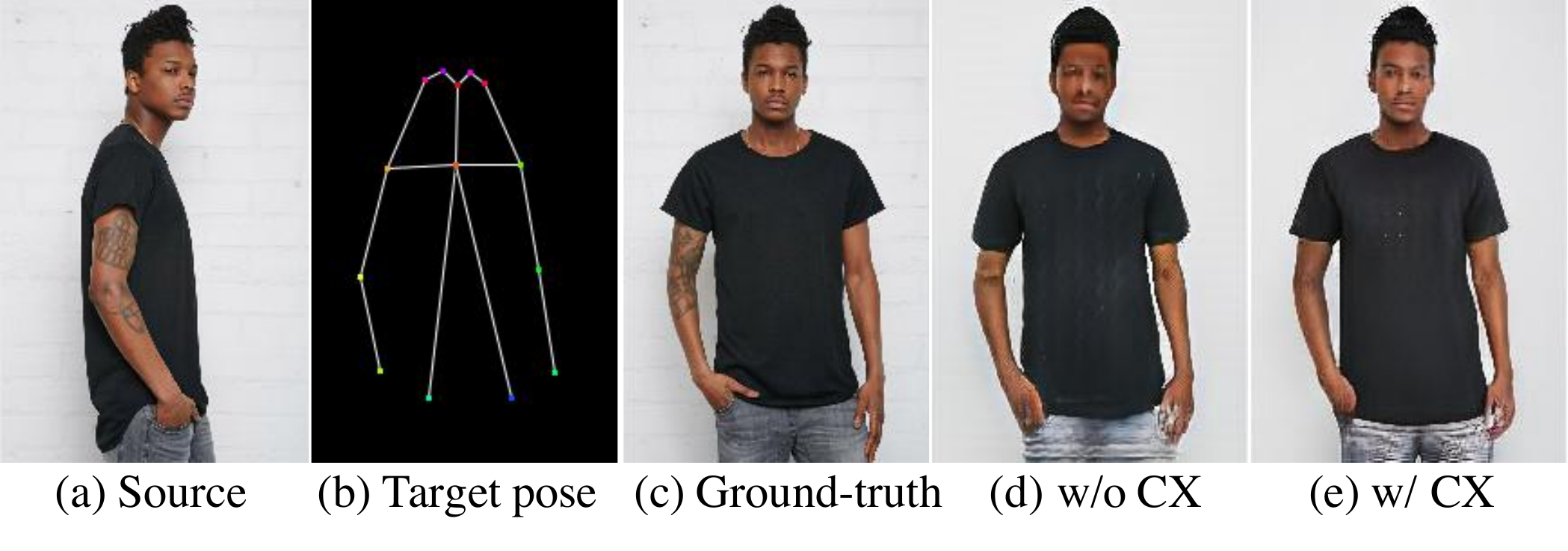}
\caption{Effects of the contextual loss.}
\label{fig:cx}
\end{figure}

{ \noindent \bf Contextual loss.}
The contextual loss proposed in ~\cite{mechrez2018contextual} is designed to measure the similarity between two non-aligned images for image transformation, which is also effective in our GAN-based person image synthesis task.
Compared with the pixel-level loss requiring pixel-to-pixel alignment, the contextual loss allows spatial deformations with respect to the target, getting less texture distortion and more reasonable outputs. We compute the contextual loss $L_{CX}$ by
\begin{equation}
\mathcal{L}_{CX} = -log(CX(\mathcal{F}^l(I_g), \mathcal{F}^l(I_t))),
\end{equation}
where $\mathcal{F}^l(I_g)$ and $\mathcal{F}^l(I_t)$ denote the feature maps extracted from layer $l=relu\{3\_2,4\_2\}$ of the pretrained VGG19 network for images $I_g$ and $I_t$, respectively, and $CX$ denotes the similarity metric between matched features, considering both the semantic meaning of pixels and the context of the entire image. More details can be found in ~\cite{mechrez2018contextual}. We show the effects of $\mathcal{L}_{CX}$ in Figure~\ref{fig:cx}, which enables the network to better preserve details with less distortion.

{ \noindent \bf Implementation details. }
Our method is implemented in PyTorch using two NVIDIA Tesla-V100 GPUs with 16GB memory. With the human parser ~\cite{badrinarayanan2017segnet}, we acquire the semantic map of person image and merge original labels defined in ~\cite{gong2017look} into $K (K=8)$ categories (i.e., background, hair, face, upper clothes, pants, skirt, arm and leg). The weights for the loss terms are set to $\lambda_{rec}$ = 2, $\lambda_{per}$ = 2, and $\lambda_{CX} = 0.02$. We adopt Adam optimizer ~\cite{kingma2014adam} with the momentum set to 0.5 to train our model for around 120k iterations. The initial learning rate is set to 0.001 and linearly decayed to 0 after 60k iterations. Following this configuration, we alternatively train the generator and two discriminators.

\section{Experimental Results}
In this section, we verify the effectiveness of the proposed network for attributes-guided person image synthesis tasks (pose transfer and component attribute transfer), and illustrate its superiority over other state-of-the-art methods. Detailed results are shown in the following subsections and more are available in the supplemental materials (Supp).

{ \noindent \bf Dataset. }
We conduct experiments on the In-shop Clothes Retrieval Benchmark DeepFashion~\cite{liu2016deepfashion}, which contains a large number of person images with various appearances and poses. There are totally 52,712 images with the resolution of $256\times256$. Following the same data configuration in pose transfer~\cite{zhu2019progressive}, we randomly picked 101,966 pairs of images for training and 8,750 pairs for testing.

{ \noindent \bf Evaluation Metrics. }
Inception Score (IS)~\cite{salimans2016improved} and Structural Similarity (SSIM)~\cite{wang2004image} are two most commonly-used evaluation metrics in the person image synthesis task, which were firstly used in PG\textsuperscript{2}~\cite{ma2017pose}. Later, Siarohin et al.~\cite{siarohin2018deformable} introduced Detection Score (DS) to measure whether the person can be detected in the image. However, IS and DS only rely on an output image to judge the quality in itself and ignore its consistency with conditional images. Here, we introduce a new metric called contextual (CX) score, which is proposed for image transformation~\cite{mechrez2018contextual} and uses the cosine distance between deep features to measure the similarity of two non-aligned images, ignoring the spatial position of the features.
CX is able to explicitly assess the texture coherence between two images and it is suitable for our task to measure the appearance consistency between the generated image and source image (target image), recording as CX-GS (CX-GT). Except for these computed metrics, we also perform the user study to assess the realness of synthesized images by human.

\subsection{Pose transfer}
\subsubsection{Person image synthesis in arbitrary poses}

Pose is one of the most essential human attributes and our experiments verify the effectiveness of our model in pose-controlled person image synthesis. Given the same source person image and several poses extracted from person images in the test set, our model can generate natural and realistic results even when the target poses are drastically different from the source in scale, viewpoints, etc. We show some results of our method in Figure~\ref{fig:res_pose} and more are available in Supp.

\begin{figure}
\setlength{\abovecaptionskip}{0cm}
\setlength{\belowcaptionskip}{-0.4cm}
\includegraphics[width=1\linewidth]{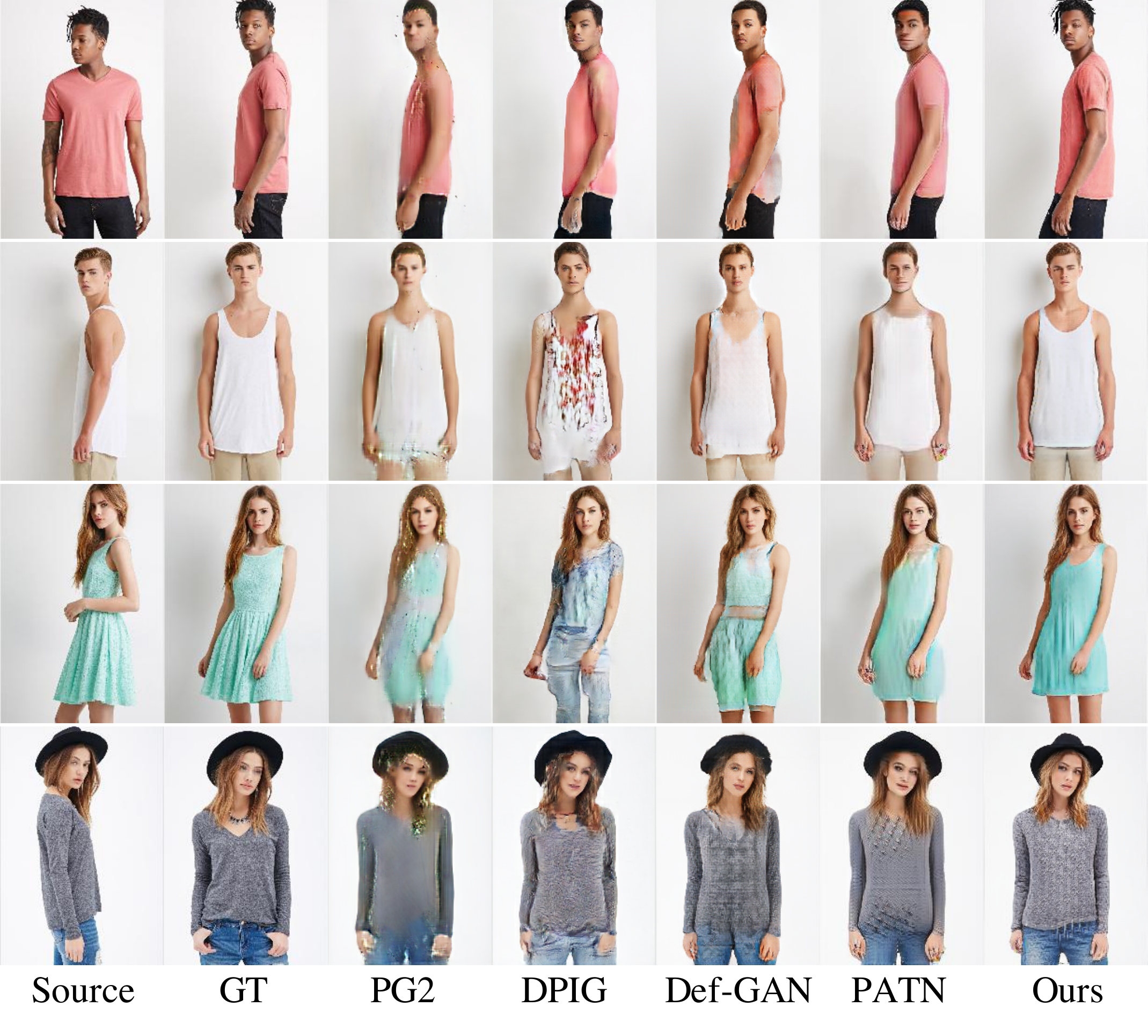}
\caption{Qualitative comparison with state-of-the-art methods. }
\label{fig:comparison}
\end{figure}

\subsubsection{Comparison with state-of-the-art methods}
For pose transfer, we evaluate our proposed method with both qualitative and quantitative comparisons.

{ \noindent \bf Qualitative comparison.} In Figure~\ref{fig:comparison}, we compare the synthesis results of our method with four state-of-the-art pose transfer methods: PG\textsuperscript{2}~\cite{ma2017pose} , DPIG~\cite{ma2018disentangled}, Def-GAN~\cite{siarohin2018deformable} and PATN~\cite{zhu2019progressive}. All the results of these methods are obtained by directly using the source codes and trained models released by authors. As we can see, our method produced more realistic results in both global structures and detailed textures. The facial identity is better preserved and even detailed muscles and clothing wrinkles are successfully synthesized. More results can be found in Supp.

{ \noindent \bf Quantitative  comparison.} In Table~\ref{table:comparison}, we show the quantitative comparison with abundant metrics described before. Since the data split information in experiments of ~\cite{ma2017pose,ma2018disentangled,siarohin2018deformable} is not given, we download their pre-trained models and evaluate their performance on our test set. Although it is inevitable that testing images may be contained in their training samples, our method still outperforms them in most metrics. The results show that our method generates not only more realistic details with the highest IS value, but also more similar and natural textures with respect to the source image and target image, respectively (lowest CX-GS and CX-GT values). Furthermore, our method has the highest confidence for person detection with the best DS value. For SSIM, we observe that when the value of IS increases, this metric slightly decreases, meaning the sharper images may have lower SSIM, which also has been observed in other methods~\cite{ma2017pose,ma2018disentangled}.

\begin{table}[]
\setlength{\abovecaptionskip}{0.1cm}
\setlength{\belowcaptionskip}{-0.2cm}

\begin{tabular*}{0.485\textwidth}{c @{\extracolsep{\fill}} |c|c|c|c|c}
\hline
Model   & IS$\uparrow$    & SSIM$\uparrow$  & DS$\uparrow$    & CX-GS$\downarrow$ & CX-GT$\downarrow$ \\
\hline\hline
PG\textsuperscript{2}     & 3.202 & 0.773 & 0.943 & 2.854 & 2.795 \\
DPIG    & 3.323 & 0.745 & 0.969 & 2.761 & 2.753 \\
Def-GAN & 3.265 & 0.770 & 0.973 & 2.751 & 2.713 \\
PATN    & 3.209 & \bf 0.774 & 0.976 & 2.628 & 2.604 \\
Ours    & \bf 3.364 & 0.772 & \bf 0.984 & \bf 2.474 & \bf 2.474\\
\hline
\end{tabular*}

\caption{Quantitative comparison with state-of-the-art methods on DeepFashion. }
\label{table:comparison}
\end{table}

\begin{table}[]
\setlength{\abovecaptionskip}{0.1cm}
\setlength{\belowcaptionskip}{-0.6cm}
\begin{center}
\begin{tabular*}{0.485\textwidth}{c @{\extracolsep{\fill}} |c|c|c|c|c}
\hline
Indicator   & PG\textsuperscript{2} &  DPIG & Def-GAN & PATN & Ours\\
\hline\hline
R2G     & 9.2 & - & 12.42 & 19.14 & \bf 23.49 \\
G2R    & 14.9 & - & 24.61 & 31.78 & \bf 38.67 \\
Prefer & 1.61 & 1.35 & 16.23 & 7.26 & \bf 73.55 \\
\hline
\end{tabular*}
\caption{Results of the user study $(\%)$. R2G means the percentage of real images rated as generated w.r.t. all real images. G2R means the percentage of generated images rated as real w.r.t. all generated images. The user preference of the most realistic images w.r.t. source persons is shown in the last row.}
\label{table:user}
\end{center}
\end{table}

\begin{figure*}
\setlength{\abovecaptionskip}{0cm}
\setlength{\belowcaptionskip}{-0.2cm}
\includegraphics[width=1\linewidth]{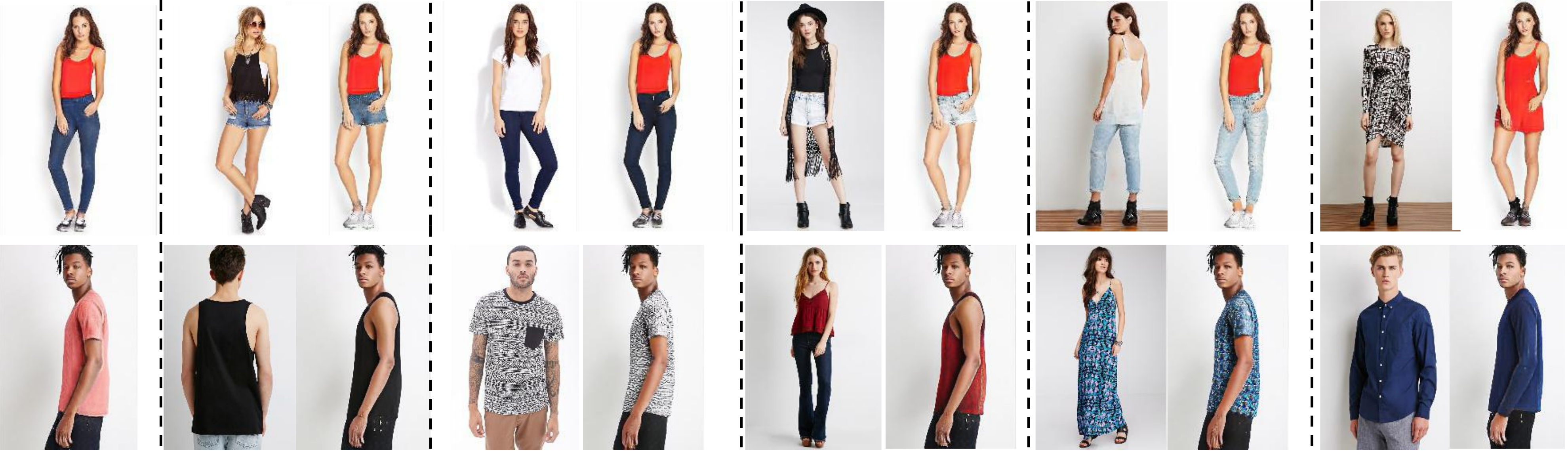}
\caption{Results of synthesizing person images with controllable component attributes. We show original person images in the first column and the images in the right are synthesized results whose pants (the first row) or upper clothes (the second row) are changed with corresponding source images in the left.}
\label{fig:mixing}
\end{figure*}

{ \noindent \bf User study.}
We conduct a user study to assess the realness and faithfulness of the generated images and compare the performance of our method with four pose transfer techniques. For the realness, participants are asked to judge whether a given image is real or fake within a second. Following the protocol of ~\cite{ma2017pose,siarohin2018deformable,zhu2019progressive}, we randomly selected 55 real images and 55 generated images, first 10 of which are used for warming up and the remaining 100 images are used for evaluation. For the faithfulness, participants are shown a source image and 5 transferred outputs, and they are asked to select the most natural and reasonable image with respect to the source person image. We show 30 comparisons to each participant and finally 40 responses are collected per experiment. The results in Table~\ref{table:user} further validate that our generated images are more realistic, natural and faithful. It is worth noting that there is a significant quality boost of synthesis results obtained by our approach compared with other methods, where over 70$\%$ of our results are selected as the most realistic one.

\subsection{Component Attribute Transfer}
Our method also achieves controllable person image synthesis with user-specific component attributes, which can be provided by multiple source person images. For example, given 3 source person images with different component attributes, we can automatically synthesize the target image with the basic appearance of person 1, the upper clothes of person 2 and the pants of person 3. This also provides a powerful tool for editing component-level human attributes, such as pants to dress, T-shirt to waistcoat, and head of man to woman.

By encoding the source person images to decomposed component codes and recombining their codes to construct the full style code, our method can synthesize the target image with desired attributes. In Figure~\ref{fig:mixing}, we edit the upper clothes or pants of target images by using additional source person images to provide desired attributes. Our method generates natural images with new attributes introduced harmoniously while preserving the textures of remaining components.


{ \noindent \bf Style Interpolation. }
Using our Attribute-Decomposed GAN, we can travel along the manifold of all component attributes of the person in a given image, thus synthesizing an animation from one attribute to another. Take for example the codes of upper clothes from person1 and person2 $(C_{uc1}$ and $C_{uc2})$, we define their mixing result as
\begin{equation}
C_{mix} = \beta C_{uc1}+ (1-\beta) C_{uc2},
\end{equation}
where $\beta \in (0,1)$ and $\beta$ decreases from 1 to 0 in specific steps. Results of style interpolation are available in Supp.

\begin{figure}
\setlength{\abovecaptionskip}{0cm}
\setlength{\belowcaptionskip}{-0.3cm}
\includegraphics[width=1\linewidth]{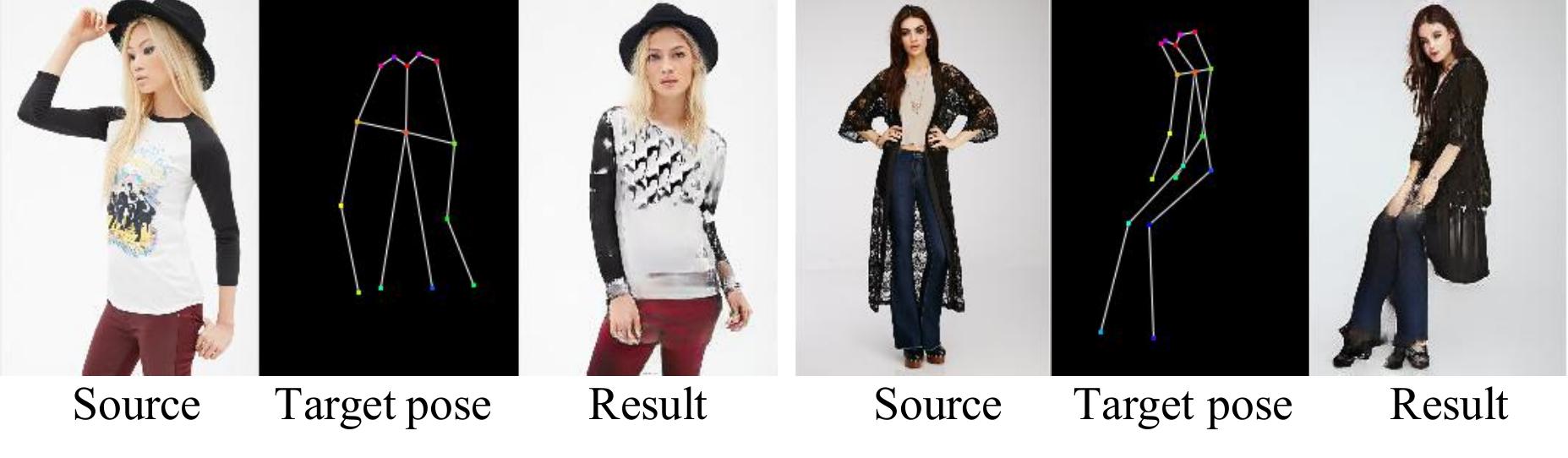}
\caption{Failure cases caused by component or pose attributes that extremely bias the manifold built upon training data.}
\label{fig:failure}
\end{figure}


\subsection{Failure cases}
Although impressive results can be obtained by our method in most cases, it fails to synthesize images with pose and component attributes that extremely bias the manifold built upon the training data. The model constructs a complex manifold that is constituted of various pose and component attributes of person images, and we can travel along the manifold from one attribute to another. Thus, valid synthesis results are actually the mixtures of seen ones via the interpolation operation. As shown in Figure~\ref{fig:failure}, the specific carton pattern in T-shirt of a woman fails to be interpolated with seen ones and the person in a rare pose cannot be synthesized seamlessly.

\section{Conclusion}
In this paper, we presented a novel Attribute-Decomposed GAN for controllable person image synthesis, which allows flexible and continuous control of human attributes. Our method introduces a new generator architecture which embeds the source person image into the latent space as a series of decomposed component codes and recombines these codes in a specific order to construct the full style code. Experimental results demonstrated that this decomposition strategy enables not only more realistic images for output but also flexible user control of component attributes. We also believed that our solution using the off-the-shelf human parser to automatically separate component attributes from the entire person image could inspire future researches with insufficient data annotation. Furthermore, our method is not only well suited to generate person images but also can be potentially adapted to other image synthesis tasks.

\section*{Acknowledgements}

This work was supported by National Natural Science Foundation of China (Grant No.: 61672043 and 61672056), Beijing Nova Program of Science and Technology (Grant No.: Z191100001119077), Key Laboratory of Science, Technology and Standard in Press Industry (Key Laboratory of Intelligent Press Media Technology).

{\small
\bibliographystyle{ieee_fullname}
\bibliography{egbib}
}

\end{document}